\documentclass[runningheads]{llncs}

\usepackage{eccv}

\usepackage{eccvabbrv}
\usepackage{booktabs,multirow,threeparttable}
\usepackage{graphicx}
\usepackage[table]{xcolor}
\usepackage{cuted}   
\usepackage{graphicx}
\usepackage{booktabs}
\usepackage{tabularx}

\usepackage[accsupp]{axessibility}  

\usepackage[pagebackref,breaklinks,colorlinks,citecolor=eccvblue]{hyperref}

\usepackage{orcidlink}
\usepackage{caption}
\begin{document}

\title{DAP: A Discrete-token Autoregressive Planner for Autonomous Driving} 

\titlerunning{DAP}

\author{Bowen Yer\inst{1,3}\and
Bin Zhang\inst{1}\and
Qiao Sun\inst{1} \and
Hang Zhao\inst{1,2}}
\authorrunning{B. Ye~Author et al.}

\institute{Shanghai Qi Zhi Institute \and
IIIS, Tsinghua University \and
Shanghai Jiaotong University
}

\maketitle
\begin{figure}[h]
    \centering
    \includegraphics[width=0.94\textwidth]{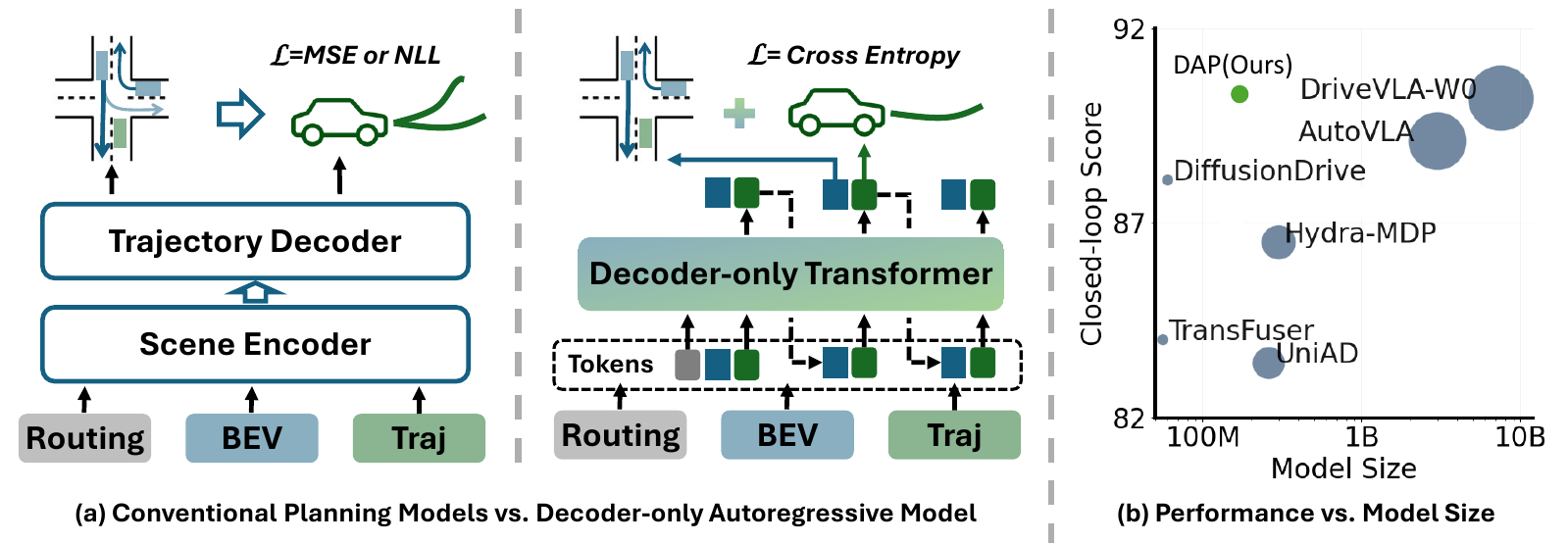}
    \caption{(a) Planning architectures: \textbf{Left} non-autoregressive direct mapping; \textbf{Right} \textbf{DAP} (ours), a discrete-token autoregressive planner that jointly forecasts environment and ego trajectories for aligned supervision and robust rollouts. (b) Performance vs.\ model size: DAP is parameter-efficient while remaining competitive with state-of-the-art methods.}
    \label{fig:shown}
    \vspace{-1.7em}
\end{figure}

\begin{abstract}
  Gaining sustainable performance improvement with scaling data and model budget remains a pivotal yet unresolved challenge in autonomous driving. While autoregressive models exhibited promising data-scaling efficiency in planning tasks, predicting ego trajectories alone suffers sparse supervision and weakly constrains how scene evolution should shape ego motion. Therefore, we introduce DAP, a discrete-token autoregressive planner that jointly forecasts BEV semantics and ego trajectories, thereby enforcing comprehensive representation learning and allowing predicted dynamics to directly condition ego motion. In addition, we incorporate a reinforcement-learning-based fine-tuning, which preserves supervised behavior cloning priors while injecting reward-guided improvements. Despite a compact 120M parameter budget, DAP achieves state-of-the-art performance on open-loop metrics and delivers competitive closed-loop results on the NAVSIM benchmark. Overall, the fully discrete-token autoregressive formulation operating on both rasterized BEV and ego actions provides a compact yet scalable planning paradigm for autonomous driving.
  \keywords{Autonomous driving \and Discrete token planner \and Autoregressive model}
\end{abstract}
\section{Introduction}
\label{sec:intro}
Research on autonomous driving planning can be categorized, from a temporal modeling perspective, into two paradigms: \emph{autoregressive (AR)} approaches that causally decode the ego’s actions one step at a time  ~\cite{11093346,feng2025artemisautoregressiveendtoendtrajectory,huang2025drivegptscalingautoregressivebehavior} and \emph{non-AR approaches} that generate the entire future trajectory in a single forward pass. The latter encompasses methods ranging from end-to-end predictive methods ~\cite{jiao2025evadriveevolutionaryadversarialpolicy, li2024hydra} that directly map sensor data to ego actions following certain command queries, to diffusion-based generative methods ~\cite{jiang2025transdiffuserdiversetrajectorygeneration, liao2025diffusiondrive} that model the distribution of ego actions conditioned on sensor data and generate planning trajectories via sampling and iterative refinement. Although non-AR methods have been extensively studied, AR approaches are gaining increasing attention in recent research~\cite{seff2023motionlm, xu2024drivegpt4interpretableendtoendautonomous}, mainly because of their superior potential for scaling up.

Recent results on large language models show that \emph{decoder-only} Transformers trained as next-token predictors over discrete text tokens scale efficiently with data, model size, and compute budget, which exhibits a predictable power-law trend ~\cite{hoffmann2022trainingcomputeoptimallargelanguage,kaplan2020scalinglawsneurallanguage}. Meanwhile, the scaling laws of autonomous driving have also been reported, with studies showing that both open- and \emph{closed-loop} metrics improve with training compute and the compute-optimal balances between model and data size have also been characterized~\cite{baniodeh2025scaling}. Building on this, DriveVLA-W0 argues that, under a comparable resource budget, \emph{autoregressive} planners scale more efficiently than other query-based or diffusion-based counterparts~\cite{li2025drivevla}. Guided by this evidence, we cast motion forecasting and planning as a discrete-token sequence modeling task and address it using a \textbf{decoder-only} Transformer with a pre-defined tokenization scheme, to thereby leverage the favorable scaling law of such architecture and ground progress in rigorous closed-loop evaluation and a compute-centric development roadmap.

Nevertheless, scaling alone does not remedy the supervision sparsity, which is an issue that limits the performance of previous models without explicit world modeling capacity. To fill this gap within our framework, we incorporate a world-modeling–style objective~\cite{ICLR20256aa49679,cen2025worldvlaautoregressiveactionworld,li2025unifiedvideoactionmodel}: the model \emph{jointly} predicts future semantic BEV representations of the environment along with discrete $\kappa$–$a$ (curvature and acceleration) action tokens of the ego at every step. By jointly forecasting BEV semantics and future trajectories, we provide dense spatio-temporal supervision. This couples scene evolution with ego motion in the latent state and improves multi-step credit assignment beyond sparse waypoint labels.

As illustrated in Figure~\ref{fig:shown}(a) right, we first tokenize the historical BEV using a VQ-VAE~\cite{NIPS2017_7a98af17}, yielding discrete environment tokens. Together with discretized past action tokens, these tokens are fed into a decoder-only autoregressive Transformer to generate future token sequences. At each timestep, the decoder jointly predicts (i) semantic BEV tokens capturing near-future scene evolution and (ii) $\kappa$--$a$ trajectory tokens governing ego motion, thereby coupling scene forecasting with motion generation under dense, spatio-temporally aligned supervision. This discrete token scheme stabilizes interactions between modules and enables efficient token-level rollouts at inference. In contrast to the left-hand baseline that maps history to a future trajectory in a single forward pass, DAP forecasts the evolving environment and ego motion in an interleaved autoregressive manner, improving robustness under closed-loop execution. Notably, as shown in Figure~\ref{fig:shown}(b), DAP remains highly parameter-efficient, achieving competitive (and often superior) performance to state-of-the-art methods despite using substantially fewer model parameters.

Following this design, we find that pure imitation learning (IL), though tends to fit ground-truth trajectories well, only yields weak coupling between ego planning and predicted scene evolution. To address this, we adopt SAC-BC (soft-actor-critic plus behavior-cloning)~\cite{lu2023imitation} fine-tuning, which preserves behavior-cloning priors while reinforcing them by leveraging reward signals, so that future environment forecasting can more directly shape trajectory generation. It is proven that our model remains compact, achieves state-of-the-art results on open-loop evaluation, and delivers strong closed-loop performance on NavSim benchmark, despite the small parameter count. Our main contributions are as follows.
\begin{itemize}
  \item \textbf{Decoder-only autoregressive MoE with discrete tokens.} We propose \textbf{DAP}, a \underline{\textbf{d}}iscrete-token \underline{\textbf{a}}uto-regressive \underline{\textbf{p}}lanner with decoder-only Transformer architecture and sparse MoE layers. The DAP generates \emph{discrete} scene and trajectory tokens autoregressively, yielding a simple interface and efficient decoding.
  \item \textbf{Joint environment--trajectory forecasting.} DAP jointly predicts future semantic BEV and $\kappa$--$a$ trajectory tokens, providing dense, spatio-temporally aligned supervision that tightly couples scene understanding with motion generation. At each time step, BEV tokens are generated in parallel with bidirectional self-attention, while trajectory tokens attend to BEV tokens via causal attention, preserving temporal causality in motion generation.

  \item \textbf{SAC-BC fine-tuning beyond pure IL.} SAC-BC surpasses pure IL while preserving architectural simplicity, and it strengthens the coupling between predicted environment states and generated ego trajectories.
  \item \textbf{Compact yet strong performance.} With a small parameter budget, the model achieves \emph{state-of-the-art} open-loop results and strong closed-loop results on NavSim.
\end{itemize}

\section{Related Works}
\label{sec:related}
\subsection{End-to-end Models in Autonomous Driving}
End-to-end trajectory planning in autonomous driving has evolved from early perception–prediction–planning stacks to unified models that learn plans directly from multimodal inputs~\cite{hu2023planning,chen2024vadv2,sun2024generalizingmotionplannersmixture}. Within this landscape, two complementary design ideas have emerged. The first treats planning as \emph{autoregressive} sequence modeling with Transformers, generating future states or trajectories step by step and, in recent large-model variants, operating on \emph{discrete tokens} with GPT-style decoders~\cite{11093346,feng2025artemisautoregressiveendtoendtrajectory,li2025drivevla,huang2025drivegptscalingautoregressivebehavior}. The second embraces \emph{world modeling}, jointly forecasting scene evolution and ego motion to provide dense, time-aligned supervision that strengthens the coupling between environment prediction and plan generation~\cite{ICLR20256aa49679,cen2025worldvlaautoregressiveactionworld,li2025unifiedvideoactionmodel}. These ideas are not opposed; recent systems such as DriveVLA-W0~\cite{li2025drivevla} and UMGen~\cite{wu2025generatingmultimodaldrivingscenes} exemplify their combination by using discrete-token autoregression within a world-modeling context. However, these works target on pixel-level world models have to adopt heavy pretrained vision or vision-language models as their backbone. In contrast, our method focuses on world modeling in compact BEV-latent space, where discrete BEV-semantic tokens are predicted in a interleaved manner along with ego motion tokens. The advantage is that our method makes the system lighter and simpler to train, avoids tedious image rollouts and remains RL-friendly, while still providing planning-centric alignment and dense supervision.

\subsection{Reinforcement learning for trajectory planning}
Pure imitation learning (IL) often overfits to demonstrations, yielding trajectories that closely mimic the expert while \emph{under-attending} to critical scene factors; under covariate shift or out-of-distribution (OOD) conditions, such policies are prone to compounding errors and risks of collisions. This limitation motivates the use of reinforcement learning (RL), which optimizes task-level objectives beyond trajectory matching. Adversarial or preference-driven policy optimization, such as APO~\cite{jiao2025evadriveevolutionaryadversarialpolicy} and earlier adversarial formulations~\cite{cheng2023adversarial}, as well as GRPO-style updates~\cite{zhou2025autovla} with precedents in general-purpose RL training~\cite{shao2024deepseekmath}, have been adopted to improve closed-loop behavior. IL+RL hybrids combine imitation losses with reward-driven objectives to balance stability and performance. Representative examples include \emph{SAC-BC}~\cite{lu2023imitation}, which augments behavior cloning with soft actor–critic updates, and ReCogDrive~\cite{li2025recogdrivereinforcedcognitiveframework}’s joint optimization of RL and BC terms (\( \mathcal{L}_{\mathrm{RL}}+\mathcal{L}_{\mathrm{BC}} \)) to couple language-conditioned reasoning with planning. 


\subsection{Trajectory Post-tuning}
\label{sec:post_tuning}
Planner outputs can be processed with \emph{post-tuning} for safety and comfort, using rule-based layers or lightweight optimizers. Representative approaches include safety shields that detect violations and trigger conservative fallbacks~\cite{9811576}, and smoothing via convexified collision/kinematic constraints ~\cite{schulman2014motion}. We follow this practice with a minimal post-tuning module that attenuates lateral jitter and jerk, serving as a constraint-aware polish rather than a second planner~\cite{shao2023safety}.

\section{Methodology}
\label{sec:method}

\begin{figure*}[!t]
  \centering
  \includegraphics[width=0.85\textwidth,trim={0 1mm 0 1mm},clip]{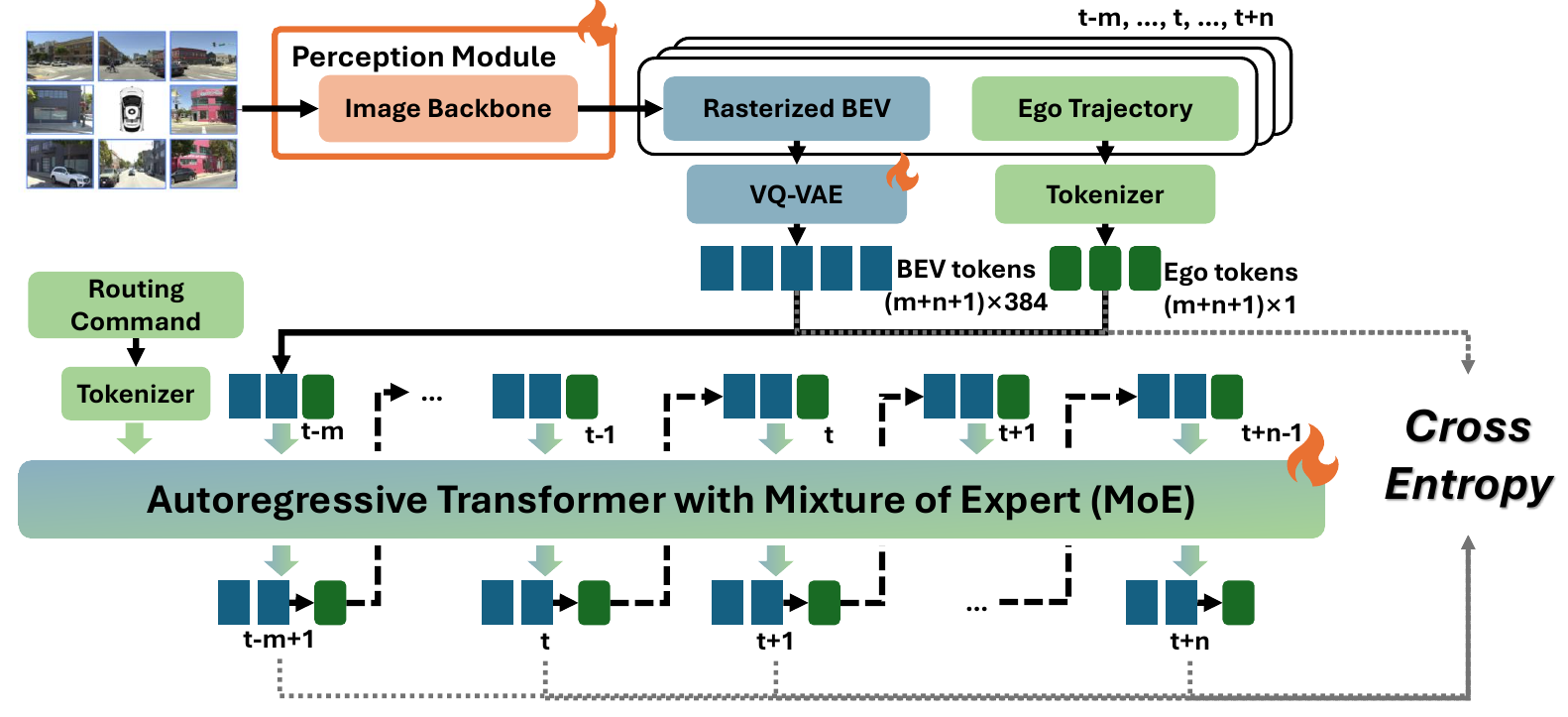}
  \caption{Overall architecture of \textbf{DAP}. Historical multi-modal inputs are tokenized (VQVAE for BEV, $\kappa$--$a$ discretization for trajectory), then a decoder-only autoregressive Transformer with sparse MoE jointly predicts future BEV and ego trajectory tokens. The joint forecasting provides dense, time-aligned signals that couple scene evolution with motion generation.}

  \label{fig:model}
\end{figure*}

\begin{figure}
    \centering
    \includegraphics[width=0.9\linewidth]{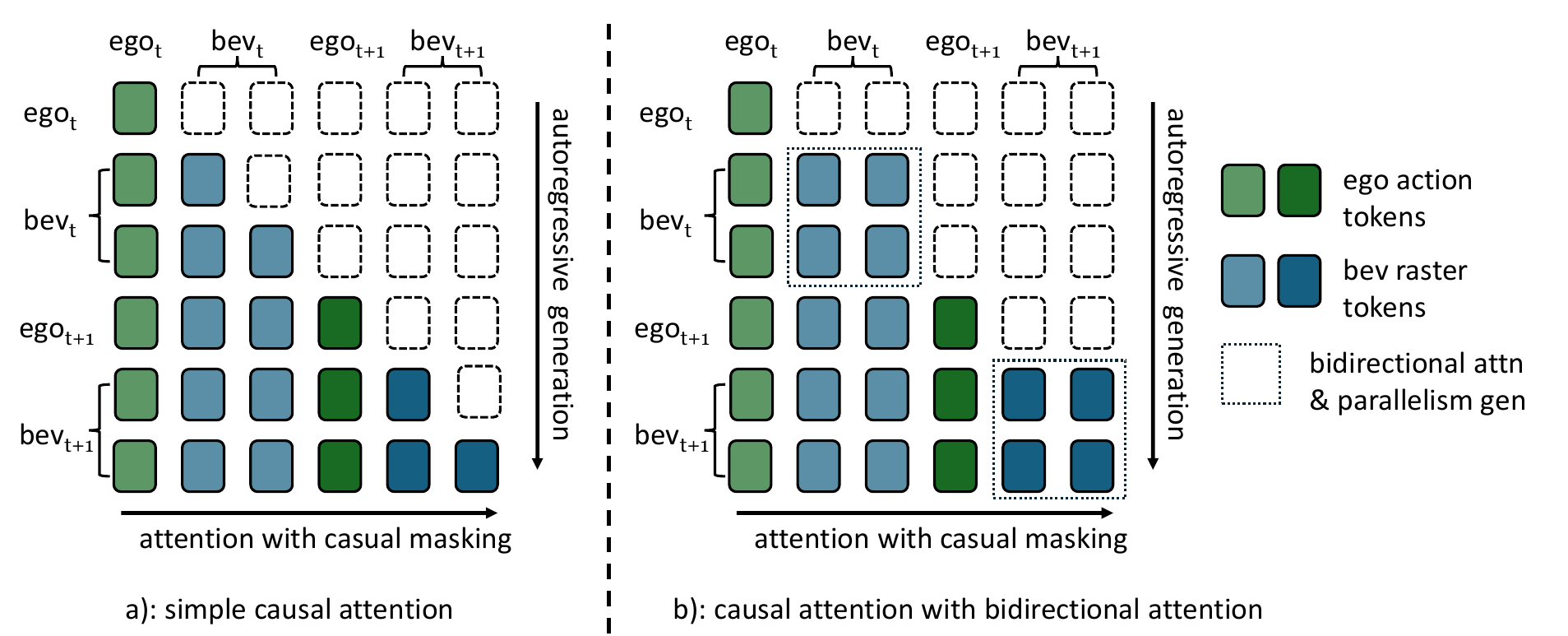}
    \caption{A comparison between (a) standard causal attention with token-by-token generation and (b) our proposed bidirectional attention mechanism applied at each BEV token generation step.}
    \label{fig:attention}
\end{figure}

In this section, we propose \emph{DAP}, in which scene understanding is aligned with motion generation through three components: 
(i) A \emph{discrete-token}, decoder-only autoregressive transformer with sparse MoE routing jointly forecasts future semantic BEV states and $\kappa$--$a$ trajectory tokens, providing dense, spatio-temporally aligned supervision.
(ii) A SAC-BC-based \emph{offline} RL stage fine-tunes the planner beyond pure imitation learning, leveraging reward signals to strengthen the coupling between predicted environment evolution and trajectory decisions while preserving architectural simplicity.
(iii) A lightweight \emph{trajectory post-tuning} step applies rule-based checks to improve ride comfort and reduce lateral deviation without modifying the planner or its discrete interface.
Together, these components yield a compact and efficient pipeline that maintains end-to-end inference while enhancing closed-loop robustness.

\subsection{Model Structure}
We adopt a planning model that conditions on a high-level command, multi-view camera observations, and ego history to predict future BEV-semantic and trajectory tokens. A perception module aggregates the image streams into BEV features, which are discretized into environment tokens via a VQ-VAE. A decoder-only autoregressive Transformer with sparse MoE then jointly decodes future BEV and trajectory tokens over the planning horizon, conditioned on the command, environment tokens, and past trajectory tokens (Fig.~\ref{fig:model}). The sparse MoE architecture increases effective capacity while enabling expert specialization over diverse traffic patterns and scene configurations, improving robustness without prohibitive inference overhead. Overall, this fully discrete, time-aligned representation couples scene evolution and ego motion within a single sequence, supporting dense supervision and favorable scaling.

To further accelerate token generation, we introduce \emph{bidirectional} (intra-step) attention within each BEV token-generation step. Specifically, all BEV tokens within the same timestep can attend bidirectionally to one another and generate in parallel, rather than being constrained by a causal mask and sequential generation. This reduces the number of autoregressive iterations, yielding a substantial speedup without hindering performance. The attention scheme is illustrated in Fig.~\ref{fig:attention}. In our experiments, DAP predicts the future 8-step trajectory together with all the BEV tokens in approximately \textbf{100\,ms} per sample.
\vspace{-1em}
\paragraph{a) Input Tokenization.}
We discretize three modalities with respective quantization schemes.

\noindent\textbf{Command:}
We treat the routing command as a $C$-class categorical variable, and convert each command $c \in \{1,\ldots,C\}$ into a one-hot vector $\mathbf{o}\in\{0,1\}^C$.

\noindent\textbf{BEV feature:}
We fuse multi-view observations into a semantic BEV feature map $\mathbf{F}\in\mathbb{R}^{H\times W\times D}$,
and quantize it using a trained VQ-VAE. The encoder $E$ produces a latent grid
$\mathbf{Z}=E(\mathbf{F})\in\mathbb{R}^{h\times w\times d}$. Given a codebook
$\mathcal{E}=\{\mathbf{e}_k\}_{k=1}^K,\ \mathbf{e}_k\in\mathbb{R}^d$, each latent vector
$\mathbf{z}_{i,j}$ is vector-quantized by nearest neighbor:
\begin{equation}
    k^\star(i,j)=\arg\min_{k\in\{1,\ldots,K\}}\left\lVert \mathbf{z}_{i,j}-\mathbf{e}_k\right\rVert_2^2.
\end{equation}
The BEV tokens are extracted as the flattened indices $\{k^\star(i,j)\}$ sequence over the latent grid.

\noindent\textbf{Ego states.}
Given ego poses $\{(x_t,y_t,\psi_t)\}_{t=0}^{T-1}$ sampled with intervals $\Delta t_t$,
we convert positions and yaw into curvature--acceleration pairs $(\kappa_t,a_t)$.
Let $\mathbf{p}_t=(x_t,y_t)$ and $\mathrm{wrap}(\cdot)$ map angles to $(-\pi,\pi]$.
We estimate the translational speed by finite differences,
$s_t=\lVert \mathbf{p}_{t+1}-\mathbf{p}_t\rVert_2/\Delta t_t$, and use a trapezoidal smoothing to obtain node-wise speeds $\{v_t\}$.
Then,
\begin{subequations}\label{eq:ka}
\begin{align}
    a_t &= \frac{v_{t+1}-v_t}{\Delta t_t},\quad v^{\mathrm{mid}}_t=\tfrac{1}{2}(v_t+v_{t+1})\\
    \kappa_t &= \frac{\mathrm{wrap}(\psi_{t+1}-\psi_t)}{\Delta t_t\,\max(v^{\mathrm{mid}}_t,\varepsilon)},
\end{align}
\end{subequations}
where $\varepsilon>0$ is to avoid numerical issues. We set $a_{T-1}=a_{T-2}$ and $\kappa_{T-1}=\kappa_{T-2}$ for a length-$T$ sequence.

We then discretize $(\kappa_t,a_t)$ into indices $(i^\kappa_t,i^a_t)$.
Curvature is quantized by a piecewise grid of  $K$ bins, with finer resolution around zero and coarser in the outer range.
Acceleration is quantized by a uniform grid of $A$ bins on $[a_{\min},a_{\max}]$ with step size $\delta_a$.
We optionally pack the pair into a single action token:
\begin{equation}
    \mathrm{S}_t = i^\kappa_t \cdot A + i^a_t \ \in\ \{0,\ldots, AK-1\},
\end{equation}
 
\paragraph{b) Autoregressive Planning Transformer.}
Following~\cite{wang2025unifiedvisionlanguageactionmodel}, we concatenate multimodal chunks over $H$ history steps.
The input sequence starts with a command token $C_{t^\ast}$ at the current time $t^\ast$, and for each step $t=t^\ast-H,\dots,t^\ast$ we append the BEV token block $V_t$ followed by one action token $A_t$:
\[
\mathbf{z}_{t^\ast-H:t^\ast}
=\big[\, C_{t^\ast},\; V_{t^\ast-H},\, A_{t^\ast-H},\; \ldots,\; V_{t^\ast},\, A_{t^\ast} \big],
\]
where $V_t \equiv [V_{t,1},\dots,V_{t,M}]$ contains $M$ BEV tokens at step $t$.
All tokens are mapped into a shared embedding space and fed into a decoder-only Transformer that maintains a causal state over the prefix $\mathbf{z}_{\le p}$ at each position $p$.

Given the observed context $\mathbf{z}_{\le t}$, the Transformer predicts future tokens in a causal manner across timesteps.
Crucially, within each future timestep, we generate the block of BEV tokens in parallel using bidirectional intra-step attention and then generate the action token conditioned on the new BEV tokens:
\begin{subequations}
\begin{align}
& p_\theta\!\big(V_{t+1,1:M} \mid \mathbf{z}_{\le t}\big)
= \prod_{m=1}^{M}\mathrm{softmax}\!\big(W_{\text{out}}\, h^{\text{bev}}_{t+1,m}\big), \label{eq:bev_parallel} \\[1mm]
& p_\theta\!\big(A_{t+1} \mid \mathbf{z}_{\le t}, V_{t+1,1:M}\big)
= \mathrm{softmax}\!\big(W_{\text{out}}\, h^{\text{act}}_{t+1}\big), \label{eq:act_cond_bev}
\end{align}
\end{subequations}
where $\{h^{\text{bev}}_{t+1,m}\}_{m=1}^{M}$ are hidden states computed for the BEV-token positions under a mask that is causal \emph{across} timesteps but bidirectional \emph{within} the BEV block of the same timestep, and $h^{\text{act}}_{t+1}$ denotes the hidden state used to emit $A_{t+1}$.
A shared output projection $W_{\text{out}}\!\in\!\mathbb{R}^{V_{\text{all}}\times d}$ corresponds to a unified discrete codebook of size $V_{\text{all}}$, with disjoint index ranges reserved for command, BEV, and trajectory tokens.
This unified formulation enables consistent cross-modality modeling, while the generation order (BEV first, then trajectory) ensures motion prediction explicitly conditions on the decoded near-future scene representation.

\vspace{-1em}
\paragraph{c) Training and Inference.}
Given ground-truth sequences $\mathbf{z}_{1:T}$, we optimize the model by maximizing the log-likelihood of next-step predictions.
To mitigate exposure bias, we adopt scheduled sampling, where the conditioning prefix $\tilde{\mathbf{z}}_{<t}$ interpolates between ground-truth and model-generated tokens:
\begin{equation}
\mathcal{L}_{\text{AR}}
= - \sum_{t=1}^{T} \log p_\theta(\mathbf{z}_{t} \mid \tilde{\mathbf{z}}_{<t}),\quad
\tilde{\mathbf{z}}_{<t} = (1 - p)\,\mathbf{z}_{<t} + p\,\hat{\mathbf{z}}_{<t},
\end{equation}
where $\hat{\mathbf{z}}_{<t}$ denotes tokens predicted by the model and $p$ is the sampling ratio.
We gradually increase $p$ from $0$ to $1$ during training, bridging teacher forcing and inference-time behavior.

At each future step, the model outputs logits for the BEV token block $V_t$ and the action token $A_t$.
We train with a weighted sum of cross-entropy objectives for the two modalities:
\begin{subequations}
\begin{align}
\mathcal{L}_{\text{bev}}
&= - \sum_{t=0}^{T}\sum_{m=1}^{M}
\log p_\theta\!\big(V_{t,m} \mid \tilde{\mathbf{z}}_{<t}\big), \label{eq:loss_bev_parallel}\\
\mathcal{L}_{\text{traj}}
&= - \sum_{t=0}^{T}
\log p_\theta\!\big(A_t \mid \tilde{\mathbf{z}}_{<t}, V_{t,1:M}\big), \label{eq:loss_traj}
\end{align}
\end{subequations}
where $T$ denotes the planning horizon.
The total loss is a weighted sum:
\[
\mathcal{L}_{\text{total}} = 
\lambda_{\text{traj}}\, \mathcal{L}_{\text{traj}} 
+ \lambda_{\text{bev}}\, \mathcal{L}_{\text{bev}},
\]
where $\lambda_{\text{traj}}$ and $\lambda_{\text{bev}}$ weight the trajectory and BEV losses, respectively. We use a larger $\lambda_{\text{traj}}$ to prioritize motion accuracy and a smaller $\lambda_{\text{bev}}$ to stabilize scene representation without dominating optimization. This asymmetric weighting produces a coherent BEV context that supports more accurate, physically plausible trajectories. The autoregressive design enforces temporal causality and inter-step consistency, and captures ego–scene interactions.



\begin{figure}[t]
  \centering
  \begin{minipage}[t]{0.42\linewidth}
    \centering
    \includegraphics[width=\linewidth]{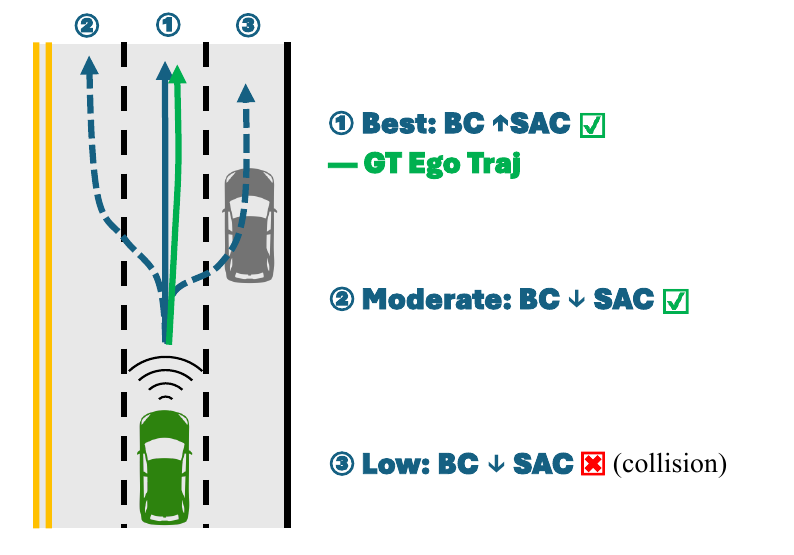}
    \caption{The necessity of RL: for the sub-optimal trajectories 2 and 3 with nearly identical BC losses, the 3rd one would yield a collision and hence get a higher RL loss.}
    \label{fig:RL}
  \end{minipage}
  \hfill
  \begin{minipage}[t]{0.54\linewidth}
    \centering
    \includegraphics[width=\linewidth]{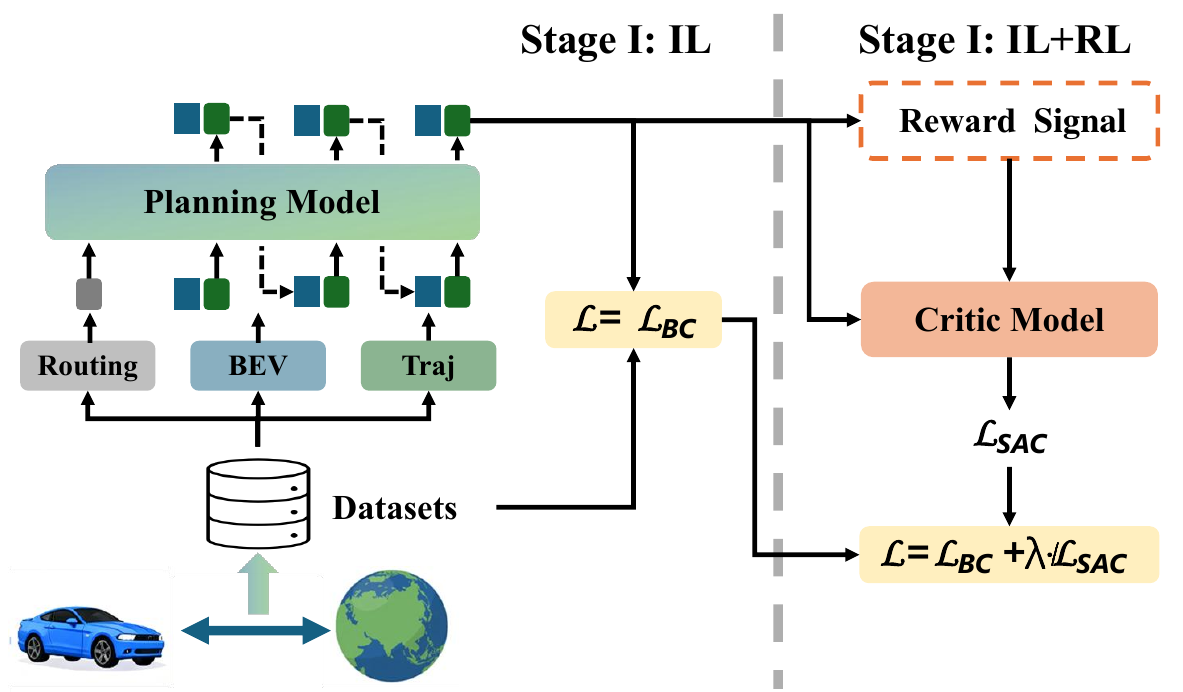}
    \caption{Two-stage training: (I) supervised pretraining with cross-entropy losses, and (II) offline SAC-BC fine-tuning that augments the policy with reward-driven adaptation while retaining behavior consistency.}
    \label{fig:training}
  \end{minipage}

\end{figure}

\subsection{Reinforcement Learning(RL)}

The supervised objective (e.g., waypoint MSE or cross-entropy over discrete tokens) treats several feasible future trajectories as loss-equivalent relative to the expert label. In the scene of Fig.~\ref{fig:RL}, “drift left”, and “drift right” can incur nearly identical surrogate loss, despite having distinct risk profiles (the left retains a larger safety buffer, while the right collides with another vehicle). This ambiguity and symmetry of the surrogate loss induce mode averaging or arbitrary mode selection, such as lateral dithering or suboptimal lane choice with no incentive for safety margins. 

The dense supervision of joint BEV and trajectory token prediction mainly reduces imitation errors, while the decision ambiguity remains. The model can still select a riskier mode that is equivalent under the surrogate loss. IL also inherits dataset biases and is brittle under covariate shift. We therefore adopt SAC-BC~\cite{lu2023imitation}, which optimizes explicit rewards for safety and comfort while regularizing toward the BC prior. This breaks the loss symmetry, offers corrective signals to avoid hazardous modes even when ego deviates from the expert manifold, and preserves the discrete autoregressive interface.

\noindent\textbf{SAC-BC with only actions.}
The trajectory token $A_t\in\{0,\ldots,A{-}1\}$ is the only action.
BEV tokens $V_{t,1:M}$ are supervised and appear only in the causal prefix
$\mathrm{ctx}_t=[\,C_{t^\ast},\,V_{t^\ast-H},A_{t^\ast-H},\ldots,V_t,A_{t-1}\,]$.
Within each step the backbone predicts $V_{t,1:M}$ and then the policy selects $A_t$ from $\pi_\phi(A_t\mid \mathrm{ctx}_t)$.

\noindent\textbf{Reward signal.}
Let $d_{\text{ctr}}(t)$ and $d_{\text{clr}}(t)$ denote the ego’s distances (in meters) at step $t$ to the lane centerline and to the nearest obstacle, respectively (measured in the BEV frame). We use bounded, scaled geometry rewards
\begin{equation}
\resizebox{0.9\columnwidth}{!}{$
r_{\text{ctr}}(t)=\big[\,1-\tfrac{d_{\text{ctr}}(t)}{\sigma_{\text{ctr}}}\,\big]_+;
r_{\text{clr}}(t)=\big[\,\tfrac{d_{\text{clr}}(t)}{\sigma_{\text{clr}}}\,\big]_+;
[x]_+\!=\!\max(x,0).$
}
\end{equation}
For comfort, we compute kinematics $v_t,a_t,\omega_t$ from $(x_t,y_t,\mathrm{yaw}_t)$ and penalize acceleration variation and angular acceleration under a low-speed mask:

\begin{equation}
r_{\text{comf}}(t)\;=\;-\Big(\lambda_{\Delta a}\,|\Delta a_t|+\lambda_{\alpha}\,|\alpha_t|\Big)\,\mathbf{1}\{|v_t|>\varepsilon_{\text{spd}}\},
\end{equation}
where $\Delta a_t=a_t-a_{t-1}$ and $\alpha_t$ is the angular acceleration derived from $\omega_t$.

The per-step reward used by SAC is
\begin{equation}
r_t \;=\;
w_{\text{ctr}}\,r_{\text{ctr}}(t)+w_{\text{clr}}\,r_{\text{clr}}(t)+w_{\text{comf}}\,r_{\text{comf}}(t).
\end{equation}

\noindent\textbf{SAC target on the prefix.} The above equation defines the SAC target used for updating the critic during prefix fine-tuning.

\begin{align}
y_t
= r_t
 + \gamma\,\mathbb{E}_{A'\sim \pi_\phi(\cdot\mid \mathrm{ctx}_{t+1})}
 \Big[\min_{i\in\{1,2\}}\,\bar Q_{\bar\theta_i}(\mathrm{ctx}_{t+1},A') - \alpha\log \pi_\phi(A'\mid \mathrm{ctx}_{t+1})\Big],\notag
\end{align}
where $A'\!\sim\!\pi_\phi(\cdot\mid \mathrm{ctx}_{t+1})$ is the next-step action used for bootstrapping, 
$\bar Q_{\bar\theta_i}$ are target critics, $\gamma\!\in\!(0,1)$ is the discount, and $\alpha\!\ge\!0$ is a fixed entropy weight.

\noindent\textbf{Critic.} We use a soft Bellman target $y_t$ (SAC) with clipped double-$Q$ and a conservative regularizer (CQL) to regress twin critics:
\begin{align}
    \mathcal{L}_{\text{critic}}
      =\tfrac{1}{2}\!\sum_{i=1}^{2}\!\big(Q_{\theta_i}(\mathrm{ctx}_t,A_t)-y_t\big)^2 +\alpha_{\text{cql}}\!\sum_{i=1}^{2}\!\Big[\log\!\sum_{a} e^{Q_{\theta_i}(\mathrm{ctx}_t,a)}-Q_{\theta_i}(\mathrm{ctx}_t,A_t)\Big]. \notag
\end{align}
where $Q_{\theta_i}$ are trainable twin critics and $\alpha_{\text{cql}}\!\ge\!0$ controls the conservative penalty that down-weights OOD actions.

\noindent\textbf{Actor.} The policy minimizes the KL divergence to the Boltzmann distribution induced by $Q$ (discrete actions use exact summation):
\begin{align}
\mathcal{L}_{\text{actor}}
= \mathbb{E}_{\mathrm{ctx}}\!\left[
  \alpha \sum_{a}\pi_\phi(a\mid\mathrm{ctx})\log \pi_\phi(a\mid\mathrm{ctx})
\right. \nonumber\left.
  - \sum_{a}\pi_\phi(a\mid\mathrm{ctx})\,\min_{i} Q_{\theta_i}(\mathrm{ctx},a)
\right].
\end{align}

\noindent\textbf{SAC loss.} The final SAC objective combines the critic regression and actor policy terms with tunable weights, as shown below.
\begin{align}
\mathcal{L}_{\text{SAC}}
&=\lambda_{critic}\,\mathcal{L}_{\text{critic}} + \lambda_{actor}\,\mathcal{L}_{\text{actor}}.
\end{align}

\noindent\textbf{Behavior cloning (value-aware).} This objective performs value-aware behavior cloning: the critic-derived advantage ($\mathrm{Adv}_t$) compares the expert action value against the policy’s value baseline, and the exponential weight ($w_t$) (AWAC-style) upweights high-advantage expert actions in the BC loss, biasing imitation toward actions that are not only likely but also value-improving.

\begin{align}
\mathrm{Adv}_t
& = \min_{i} Q_{\theta_i}(\mathrm{ctx}_t,A_t^{\text{gt}}) - \sum_{a=0}^{A-1}\pi_\phi(a\mid \mathrm{ctx}_t)\,\min_{i} Q_{\theta_i}(\mathrm{ctx}_t,a),\\
\mathcal{L}_{\text{BC}}
&=\mathbb{E}\!\left[w_t\cdot\big(-\log \pi_\phi(A_t^{\text{gt}}\mid \mathrm{ctx}_t)\big)\right],
\end{align}
where $A_t^{\text{gt}}$ is the expert action token at step $t$, $w_t = \exp\!\big(\mathrm{Adv}_t/\lambda_{\text{awac}}\big)$ is the AWAC weight with temperature $\lambda_{\text{awac}}\!>\!0$, and $\pi_\phi(\cdot\mid \mathrm{ctx}_t)$ is the discrete policy over the trajectory-token vocabulary.

\noindent\textbf{Total objective.}
We optimize a weighted sum of the SAC and BC objectives, with $\lambda$ controlling the BC weight.
\begin{equation}
\mathcal{L}_{\text{total}}
\;=\;
\mathcal{L}_{\text{SAC}}
\;+\;
\lambda\mathcal{L}_{\text{BC}}. \notag
\end{equation}

Combined with former IL process, our two-stage training process is shown in Figure~\ref{fig:training}: (I) behavior cloning pretrains a strong perception-to-plan prior over discrete BEV and $\kappa$–$a$ tokens; (II) offline SAC-BC then \emph{breaks the loss symmetry} by optimizing explicit rewards (safety and comfort) while regularizing toward the BC policy. This shifts the learning from mere label matching to risk-aware selection, e.g., preferring the left trajectory in Fig.~\ref{fig:RL} rather then the right one.

\subsection{Trajectory Post-tuning}
The discrete $\kappa$--$a$ tokenization is compact and robust but may miss small displacements and induce abrupt token switches, occasionally manifesting as lateral zig-zag or comfort degradation. 
To mitigate these flaws, we introduce a lightweight post-tuning stage that refines the predicted trajectory using BEV lane evidence and finite-difference regularization.

\noindent\textbf{Formulation.}
Given the predicted waypoints $\{(x_t,y_t,\psi_t)\}_{t=1}^{H}$ and a lane-center likelihood map $P\!\in\![0,1]^{H\times W}$, we first obtain lane anchors $(x_t^{\text{lane}},y_t^{\text{lane}})$ by gradient ascent on $P$. 
In the local Frenet frame $(s_t,\ell_t)$, we minimize a regularized least-squares objective:
\begin{equation}\label{eq:post-lat}
\displaystyle
\min_{\Delta\boldsymbol{\ell}}
\big\|\Delta\boldsymbol{\ell}-(\ell^{\text{lane}}\!-\ell)\big\|_2^2
+w_{\ell,1}\|D_1\Delta\boldsymbol{\ell}\|_2^2
+w_{\ell,2}\|D_2\Delta\boldsymbol{\ell}\|_2^2,
\end{equation}
where $D_1,D_2$ are first/second-order finite-difference operators.  
A similar 1D smoothing is applied longitudinally:
\begin{align}
\min_{\mathbf{s}}\;&\|\mathbf{s}-\mathbf{s}_{\text{raw}}\|_2^2
+w_{s,1}\|D_1\mathbf{s}\|_2^2
+w_{s,2}\|D_2\mathbf{s}\|_2^2.
\label{eq:post-long}
\end{align}
Finally, yaw angles are recomputed from the refined $(x_t,y_t)$ and softly smoothed under a per-step rate limit.

This optimization preserves the planner’s intent while aligning the trajectory with lane geometry and enforcing local smoothness, yielding improved feasibility and ride comfort without introducing new learnable modules.

\section{Experiments}
\label{sec:experiments}

\subsection{Implementation Details}
We train on NavSim~\cite{dauner2024navsim} at 2\,Hz using a two-stage schedule: 4 epochs of behavioral cloning (BC) to stabilize representations and align with teacher trajectories, followed by joint optimization of the full objective for a total of 16 epochs. We use AdamW (lr $1{\times}10^{-4}$, weight decay $1{\times}10^{-4}$) with gradient clipping at 1.0 and $n_{\text{step}}{=}3$ rollout steps per update. Training is performed on 4$\times$A800 (80\,GB) GPUs with the effective batch size scaled with device count. For nuScenes~\cite{caesar2020nuscenes} evaluation, we train a separate model with BC only. Images are converted to BEV via the TransFuser~\cite{chitta2022transfuser} pipeline: we initialize from a pretrained $\sim$50M-parameter TransFuser encoder (ResNet-34 backbone) and further fine-tune it on our training data to fuse modalities and produce semantic BEV maps, which are then fed into the VQ-VAE and planning Transformer.

\begin{table}[t]
  \centering
  \setlength{\abovecaptionskip}{2pt}%
  \setlength{\belowcaptionskip}{2pt}%

  \caption{Main results on nuScenes. ``Avg.'' averages the first three seconds.}
  \label{tab:nuscenes}

  \begingroup
  \scriptsize
  \setlength{\tabcolsep}{4pt}%
  \renewcommand{\arraystretch}{1.05}%
  \setlength{\aboverulesep}{0pt}%
  \setlength{\belowrulesep}{0pt}%
  \setlength{\abovetopsep}{0pt}%
  \setlength{\belowbottomsep}{0pt}%
  \renewcommand\cmidrulekern{0pt}%
  \renewcommand\cmidrulesep{0pt}%
  \setlength{\heavyrulewidth}{0.6pt}%
  \setlength{\lightrulewidth}{0.3pt}%

  \begin{threeparttable}
    \begin{tabular*}{\textwidth}{@{\extracolsep{\fill}}%
      l
      c c c c
      c c c c
    @{}}
      \toprule
      \multirow{2}{*}{\bfseries Model} &
        \multicolumn{4}{c}{$\mathbf{L2_{\text{avg}}}$ (m) $\downarrow$} &
        \multicolumn{4}{c}{$\mathbf{L2_{\max}}$ (m) $\downarrow$} \\
      \cmidrule(lr){2-5}\cmidrule(lr){6-9}
      & {1s} & {2s} & {3s} & {Avg.} & {1s} & {2s} & {3s} & {Avg.} \\
      \midrule
      VAD~\cite{jiang2023vad}                   & 0.41 & 0.70 & 1.05 & 0.72 & {--} & {--} & {--} & {--} \\
      BridgeAD~\cite{zhang2025bridging}         & 0.28 & 0.55 & 0.92 & 0.58 & {--} & {--} & {--} & {--} \\
      UniAD~\cite{hu2023planning}               & 0.42 & 0.64 & 0.91 & 0.66 & 0.48 & 0.96 & 1.65 & 1.03 \\
      SparseDrive~\cite{sun2025sparsedrive}     & 0.29 & 0.58 & 0.96 & 0.61 & {--} & {--} & {--} & {--} \\
      SSR~\cite{li2024does}                     & 0.19 & 0.36 & 0.62 & 0.39 & 0.25 & 0.64 & 1.33 & 0.74 \\
      \midrule
      OpenDriveVLA~\cite{zhou2025opendrivevla}  & 0.15 & 0.31 & 0.55 & 0.33 & \textbf{0.20} & \textbf{0.58} & \underline{1.21} & \underline{0.66} \\
      EMMA$^{\ast}$~\cite{hwang2024emma}        & 0.14 & \underline{0.29} & 0.54 & 0.32 & {--} & {--} & {--} & {--} \\
      EMMA+$^{\ast}$~\cite{hwang2024emma}       & \underline{0.13} & \textbf{0.27} & \underline{0.48} & \underline{0.29} & {--} & {--} & {--} & {--} \\
      MAX-V1$^{\ast}$~\cite{yang2025less}       & 0.24 & 0.38 & 0.65 & 0.42 & 0.28 & 0.63 & 1.41 & 0.77 \\
      \textbf{Ours (DAP)}                       & \textbf{0.12} & \textbf{0.27} & \textbf{0.44} & \textbf{0.27} & \underline{0.21} & \textbf{0.50} & \textbf{1.00} & \textbf{0.57} \\
      \bottomrule
    \end{tabular*}

    \begin{tablenotes}[flushleft]\footnotesize
      \item Note: $^{\ast}$ EMMA is initialized from Google Gemini~\cite{hwang2024emma}; EMMA+ is pre-trained on Waymo's internal extra data. For MAX-V1, we show results of MiMo-VL-7B-SFT. \textbf{Bold} indicates the best result, \underline{underline} indicates the second best.
    \end{tablenotes}
  \end{threeparttable}
  \endgroup

\end{table}

\begin{table}[t]
  \centering
  \setlength{\abovecaptionskip}{2pt}%
  \setlength{\belowcaptionskip}{2pt}%

  \caption{Performance on NuPlan open-loop metrics, higher is better for OLS (\%).}
  \label{tab:nuplan}

  \begingroup
  \scriptsize
  \renewcommand{\arraystretch}{1.05}%
  \setlength{\tabcolsep}{1.4pt}

  \begin{threeparttable}
    \begin{tabular*}{\linewidth}{@{\extracolsep{\fill}} lccc ccc ccc}
      \toprule
      \multirow{2}{*}{\bfseries Method} &
        \multicolumn{3}{c}{Val4k Set} &
        \multicolumn{3}{c}{Test4k Set} &
        \multicolumn{3}{c}{Val14 Set} \\
      \cmidrule(lr){2-4}\cmidrule(lr){5-7}\cmidrule(lr){8-10}
      & 8sADE$\hspace{0.3em}\downarrow$ & 8sFDE$\hspace{0.3em}\downarrow$ & OLS$\hspace{0.3em}\uparrow$
      & 8sADE$\hspace{0.3em}\downarrow$ & 8sFDE$\hspace{0.3em}\downarrow$ & OLS$\hspace{0.3em}\uparrow$
      & 8sADE$\hspace{0.3em}\downarrow$ & 8sFDE$\hspace{0.3em}\downarrow$ & OLS$\hspace{0.3em}\uparrow$ \\
      \midrule
      PlanCNN~\cite{renz2022plant} & {--}   & {--}   & {--}   & {--}   & {--}   & {--}   & 2.468 & 5.936 & 64   \\
      PDM-Hybrid~\cite{dauner2023parting} & 2.435 & 5.202 & 84.06 & 2.618 & 5.546 & 82.04 & 2.381 & 5.068 & 84   \\
      PlanTF~\cite{cheng2024rethinking}   & 1.774 & 3.892 & 88.59 & 1.855 & 4.042 & 87.30 & 1.697 & \textbf{3.714} & 89.18 \\
      DTPP~\cite{huang2024dtpp}           & 4.196 & 9.231 & 65.15 & 4.117 & 9.181 & 64.18 & 4.088 & 8.846 & 67.33 \\
      STR2-CKS-800m~\cite{sun2024generalizingmotionplannersmixture} & 1.473 & 4.124 & 90.07 & 1.537 & 4.269 & 89.12 & 1.496 & 4.219 & 89.2 \\
      \textbf{Ours (DAP)}                 & \textbf{1.202} & \textbf{3.711} & \textbf{91.68} & \textbf{1.393} & \textbf{4.090} & \textbf{90.16} & \textbf{1.311} & 3.942 & \textbf{91.02} \\
      \bottomrule
    \end{tabular*}

    \begin{tablenotes}[flushleft]\footnotesize
      \item Note: \textbf{Bold} indicates the best result.
    \end{tablenotes}
  \end{threeparttable}
  \endgroup
\end{table}

\subsection{Open-loop Evaluation}
We conduct open-loop evaluations on nuScenes~\cite{caesar2020nuscenes} and NuPlan~\cite{caesar2021nuplan}. On nuScenes, following UniAD and ST-P3 we report $\mathrm{L2}_{\max}$ and $\mathrm{L2}_{\text{avg}}$; to avoid cross-domain leakage, our nuScenes model is trained with imitation learning only (BC stage). Under the same IL protocol, we only compare against MiMo-VL-7B-SFT rather than its MAX-V1 variants (which rely on extra RL). Our planner attains the best $\mathrm{L2}_{\max}$ and matches the top $\mathrm{L2}_{\text{avg}}$, evidencing stronger worst-case control without eroding average error. 

On NuPlan, we use the full Nuplan mini set for training and evaluate on Val4k, Test4k, and Val14 following STR2-CKS~\cite{sun2024generalizingmotionplannersmixture}. As summarized in Table~\ref{tab:nuplan}, DAP sets a new state of the art on 8s ADE and OLS across all three splits (e.g., Val4k: 1.202 ADE, 91.68\% OLS), while remaining competitive on 8s FDE—slightly above PlanTF on Test4k/Val14 yet substantially ahead of other SOTA baselines overall. The pattern is consistent: DAP improves distribution-level accuracy and reliability (ADE/OLS) and preserves strong final-step precision without resorting to task-specific tuning or additional modalities.

\begin{table}[t]
  \centering
  \setlength{\abovecaptionskip}{2pt}%
  \setlength{\belowcaptionskip}{2pt}%

  \caption{Closed-loop performance on PDMS. Higher is better ($\uparrow$). C denotes camera-only inputs, while L indicates the inclusion of LiDAR in addition to cameras.}
  \label{tab:pdms-leaderboard}

  \scriptsize
  \setlength{\tabcolsep}{3pt}%
  \renewcommand{\arraystretch}{1.05}%

  \begin{threeparttable}
    \begin{tabular*}{\linewidth}{@{\extracolsep{\fill}} l c c c c c c c c}
      \toprule
      \bfseries Method & \bfseries Ref & \bfseries Sensors &
      \bfseries NC$\uparrow$ & \bfseries DAC$\uparrow$ &
      \bfseries TTC$\uparrow$ & \bfseries C.$\uparrow$ &
      \bfseries EP$\uparrow$ & \bfseries PDMS$\uparrow$ \\
      \midrule
      Human & -- & -- & 100 & 100 & 100 & 99.9 & 87.5 & 94.8 \\
      \midrule
      UniAD~\cite{hu2023planning}           & CVPR'23   & C     & 97.8 & 91.9 & 92.9 & \textbf{100.0} & 78.8 & 83.4 \\
      TransFuser~\cite{chitta2022transfuser}   & TPAMI'23 & C + L & 97.7 & 92.8 & 92.8 & \textbf{100.0} & 79.2 & 84.0 \\
      DiffusionDrive~\cite{liao2025diffusiondrive} & CVPR'25 & C + L & 98.2 & 96.2 & 94.7 & \textbf{100.0} & 82.2 & 88.1 \\
      WoTE~\cite{li2025end}                 & ICCV'25   & C + L & 98.5 & 96.8 & 94.4 & 99.9           & 81.9 & 88.3 \\
      MeanFuser~\cite{wang2026meanfuserfastonestepmultimodal}   & CVPR'26  & C & 98.6	& 97.0	&95.0	& \textbf{100}	& 82.8	&89.0 \\
      \midrule
      AutoVLA~\cite{zhou2025autovla}        & NeurIPS'25 & C    & 98.4 & 95.6 & \textbf{98.0} & 99.9 & 81.9 & 89.1 \\
      ReCogDrive~\cite{li2025recogdrivereinforcedcognitiveframework} & arXiv'25 & C & 98.2 & 97.8 & 95.2 & 99.8 & 83.5 & 89.6 \\
      DriveVLA\mbox{-}W0$^\ast$~\cite{li2025drivevla} & ICLR'26 & C & \textbf{98.7} & \textbf{99.1} & 95.3 & 99.3 & 83.3 & \textbf{90.2} \\
      \midrule
      \rowcolor{black!7}
      \textbf{Ours (DAP)} & -- & C & 98.1 & 97.9 & 97.7 & \textbf{100.0} & \textbf{86.8} & 90.0 \\
      \bottomrule
    \end{tabular*}

    \begin{tablenotes}[flushleft]\footnotesize
      \item Note: \textbf{Bold} indicates the best result. $^\ast$ indicates the best variant of DriveVLA.
    \end{tablenotes}
  \end{threeparttable}
\end{table}

\begin{table}[t]
  \centering
  \setlength{\abovecaptionskip}{2pt}%
  \setlength{\belowcaptionskip}{2pt}%

  \caption{Closed-loop performance on EPDMS. Higher is better ($\uparrow$).}
  \label{tab:navsim_v2}

  \scriptsize
  \setlength{\tabcolsep}{3pt}%
  \renewcommand{\arraystretch}{1.05}%

  \begin{threeparttable}
    \begin{tabular*}{\linewidth}{@{\extracolsep{\fill}} l c c c c c c c c c c}
      \toprule
      \bfseries Method &
      \bfseries NC$\uparrow$ & \bfseries DAC$\uparrow$ & \bfseries DDC$\uparrow$ &
      \bfseries TLC$\uparrow$ & \bfseries EP$\uparrow$  & \bfseries TTC$\uparrow$ &
      \bfseries LK$\uparrow$ & \bfseries HC$\uparrow$  & \bfseries EC$\uparrow$  &
      \bfseries EPDMS$\uparrow$ \\
      \midrule
      Ego Status                                      & 93.1 & 77.9 & 92.7 & 99.6 & 86.0 & 91.5 & 89.4 & 98.3 & 85.4 & 64.0 \\
      TransFuser~\cite{chitta2022transfuser}               & 96.9 & 89.9 & 97.8 & 99.7 & 87.1 & 95.4 & 92.7 & 98.3 & 87.2 & 76.7 \\
      HydraMDP++~\cite{li2024hydra}                  & 97.2 & 97.5 & 99.4 & 99.6 & 83.1 & 96.5 & 94.4 & 98.2 & 70.9 & 81.4 \\
      DriveSuprem~\cite{yao2025drivesuprimprecisetrajectoryselection}          & 97.5 & 96.5 & 99.4 & 99.6 & \textbf{88.4} & 96.6 & 95.5 & 98.3 & 77.0 & 83.1 \\
      DiffusionDrive~\cite{liao2025diffusiondrive}   & 98.2 & 95.9 & 99.4 & 99.8 & 87.5 & 97.3 & 96.8 & 98.3 & 87.7 & 84.5 \\
      DriveVLA\mbox{-}W0~\cite{li2025drivevla}       & \textbf{98.5} & \textbf{99.1} & 98.0 & 99.7 & 86.4 & \textbf{98.1} & 93.2 & 97.9 & 58.9 & 86.1 \\
      MeanFuser~\cite{wang2026meanfuserfastonestepmultimodal}     &  98.3 &	97.2	& \textbf{99.6}	& \textbf{99.8}	& 87.6	& 97.4	& \textbf{97.3}	& 98.3	& \textbf{88.2}	& \textbf{89.5} \\
      \midrule
      \rowcolor{black!7}
      \textbf{Ours (DAP)}  & 97.1 & 95.2 & 98.8 & 99.7 & \textbf{88.4} & 95.7 & 95.9 & \textbf{98.9} & 70.3 & 85.6 \\
      \bottomrule
    \end{tabular*}

    \begin{tablenotes}[flushleft]\footnotesize
      \item Note: \textbf{Bold} indicates the best result.
    \end{tablenotes}
  \end{threeparttable}
\end{table}

\subsection{Closed-loop Evaluation on NavSim v1}
We follow the official NavSim~\cite{dauner2024navsim} v1 Predictive Driver Model Score (PDMS), which aggregates a series of safety compliance and weighted drivability subscores derived from a simulator.
Table~\ref{tab:pdms-leaderboard} reports the closed-loop results on PDMS. Despite being a lightweight planner with an efficient \emph{120M}-parameter backbone, \textbf{DAP} achieves a competitive PDMS of \textbf{90.0}, matching or outperforming most recent camera-only methods. Notably, \textbf{DAP} attains \textbf{perfect comfort} (C=100.0) while maintaining strong safety-related metrics (TTC=97.7, DAC=97.9), and yields the best progress score among all listed camera-only approaches (EP=86.8). In contrast, methods that slightly exceed our PDMS (e.g., DriveVLA-W0$^\ast$) typically depend on VLM backbones with billions of parameters, highlighting the favorable performance--efficiency trade-off of our approach.

\subsection{Closed-loop Evaluation on NavSim v2}
To move beyond pseudo-simulation and more stringently evaluate closed-loop driving behavior, we further benchmark \textbf{DAP} on NAVSIM v2 introduced by Cao \emph{et al.}~\cite{cao2025pseudosimulationautonomousdriving}. Compared to v1, NAVSIM v2 adds more compliance and comfort signals, including Driving Direction Compliance (DDC), Traffic Light Compliance (TLC), Lane Keeping (LK), History Comfort (HC), and Extended Comfort (EC). Overall performance is summarized by the Extended Predictive Driver Model Score (EPDMS), which multiplicatively gates progress by safety and rule compliance before aggregating weighted drivability terms. Table~\ref{tab:navsim_v2} reports closed-loop results on NAVSIM v2. \textbf{DAP} achieves an EPDMS of \textbf{85.6}, substantially improving over the ego-status baseline (64.0) and remaining competitive with strong learning-based planners. It achieves the best progress (EP=88.4) and History Comfort (HC=\textbf{98.9}). The gap to the top-performing approach is mainly attributed to extended-comfort-related terms (EC=70.3), suggesting headroom in sustaining long-horizon comfort and stability under the stricter v2 protocol.

\subsection{Qualitative Results}

\begin{figure}[!b]
    \centering
    \includegraphics[width=0.96\linewidth]{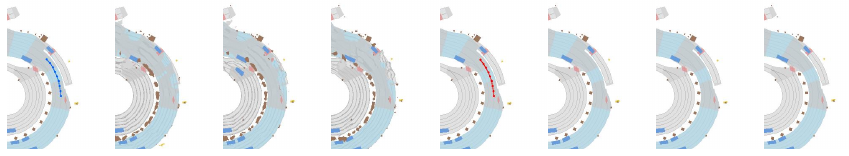}
    \caption{Qualitative results of joint planning and BEV forecasting.}
    \label{fig:results}
\end{figure}

In this section, we present qualitative visualizations in Figure~\ref{fig:results}. 
Our planner DAP jointly predicts future BEV semantics and the ego trajectory, enabling planning to explicitly account for the anticipated scene evolution. 
Each scenario is visualized with eight thumbnails: the first four summarize our predictions, including the predicted trajectory rendered on the current-frame BEV and the predicted BEV semantics for the next three horizons; the last four provide the corresponding ground truth, where the ground-truth trajectory is rendered on the same current-frame BEV followed by the ground-truth future BEV semantics. 
This layout allows an intuitive, side-by-side comparison of both motion and scene forecasting under identical visualization conditions. As shown, DAP generates trajectories that closely match the ground truth and produces high-fidelity future BEV forecasts, demonstrating strong consistency between the predicted environment representation and the planned motion.

\subsection{Ablation Study}

\begin{table}[t]
  \centering
  \setlength{\abovecaptionskip}{2pt}%
  \setlength{\belowcaptionskip}{2pt}%
  \caption{Ablation study on PDMS (NAVSIM v1). Higher is better ($\uparrow$).}
  \label{tab:ablation_pdms}

  \scriptsize
  \renewcommand{\arraystretch}{1.08}
  \setlength{\tabcolsep}{4pt}

  \begin{minipage}[t]{0.48\linewidth}
    \centering
    \textbf{Training/Objective Variants}\par\vspace{3pt}
    \begin{tabular}{p{0.66\linewidth} c}
      \toprule
      \textbf{Model} & \textbf{PDMS$\uparrow$} \\
      \midrule
      Planner (20k, traj-only) & 82.8 \\
      DAP (20k, BC)            & 84.6 \\
      DAP (20k, SACBC)         & 85.4 \\
      \bottomrule
    \end{tabular}

    \vspace{4pt}
    {\raggedright\footnotesize
    \textbf{Notes.} \{20k, 50k, 80k\} denote training set sizes. ds indicate BEV downsampling factors and C indicates codebook size.
    \par}
  \end{minipage}
  \hspace{0.02\linewidth}
  \begin{minipage}[t]{0.48\linewidth}
    \centering
    \textbf{Data Scale \& BEV Tokenization}\par\vspace{3pt}
    \begin{tabular}{p{0.66\linewidth} c}
      \toprule
      \textbf{Model} & \textbf{PDMS$\uparrow$} \\
      \midrule
      DAP (20k, ds=16, C=512)          & 85.8 \\
      DAP (20k, ds=32, C=512)         & 85.7 \\
      DAP (50k, ds=16, C=512)          & 87.7 \\
      DAP (50k, ds=32, C=512)         & 86.1 \\
      DAP (80k, ds=16, C=512)          & 88.7 \\
      DAP (80k, ds=32, C=512)         & 86.4 \\
      DAP (80k, ds=16, C=1024)       & \textbf{90.0} \\
      DAP (80k, ds=32, C=1024)      & 86.6 \\
      \bottomrule
    \end{tabular}
  \end{minipage}
\end{table}

As summarized in Table~\ref{tab:ablation_pdms}, we first ablate the supervision interface by removing the BEV prediction head and training on trajectory tokens only. This \emph{traj-only} variant yields the lowest PDMS (82.8), indicating that trajectory imitation alone fails to recover the scene-level structure required for robust planning. Joint BEV--trajectory supervision under behavioral cloning (BC) improves PDMS to 84.6, and further adopting SAC-BC brings a consistent gain (PDMS=85.4), suggesting that RL-style objectives complement imitation by enhancing closed-loop correction and safety-aware decision making.

We next examine scaling and BEV tokenization. For bev tokenization setting (ds=16, $C$=512), increasing the training data leads to monotonic improvements, with PDMS rising from 85.8 (20k) to 87.7 (50k) and 88.7 (80k), confirming that scaling data is effective for this architecture. In terms of granularity, coarser BEV tokens (ds=32) consistently underperform ds=16 at comparable scales (e.g., 86.4 vs.\ 88.7 at 80k with $C$=512), implying that excessive downsampling discards geometric details important for closed-loop robustness. Finally, enlarging the codebook to $C$=1024 substantially boosts performance under ds=16, achieving the best PDMS of \textbf{90.0} at 80k, whereas the same expansion under ds=32 yields marginal gains. Overall, the ablation supports three takeaways: (i) joint BEV and trajectory supervision is essential, (ii) SAC-BC consistently improves upon BC, and (iii) both data scaling and higher-fidelity BEV tokenization (finer ds with a larger codebook) are key to maximizing closed-loop performance.

\section{Conclusion}
\label{sec:conclu}
In this work, we argue that discrete-token, decoder-only autoregression is a promising and scalable paradigm for motion planning. We present \textbf{DAP}, a discrete-token autoregressive planner that jointly forecasts future BEV semantics and trajectory tokens, providing dense, spatio-temporally aligned supervision that tightly couples scene understanding with motion generation. Building on this foundation, we incorporate SAC-BC fine-tuning to introduce reward-driven adaptation while preserving the autoregressive decoding structure, yielding substantial improvements in closed-loop interaction. Despite a compact 120M parameter budget, DAP achieves state-of-the-art open-loop metrics and strong closed-loop performance, and our results suggest encouraging scaling trends with increased data. Finally, bidirectional intra-step attention over BEV tokens enables parallel BEV decoding and low-latency prediction, meeting the efficiency requirements for practical deployment.

%
%
\bibliographystyle{splncs04}
\bibliography{main}

@String(CVPR= {IEEE Conf. Comput. Vis. Pattern Recog.})

@String(CVPR  = {CVPR})

@inproceedings{seff2023motionlm,
  title={Motionlm: Multi-agent motion forecasting as language modeling},
  author={Seff, Ari and Cera, Brian and Chen, Dian and Ng, Mason and Zhou, Aurick and Nayakanti, Nigamaa and Refaat, Khaled S and Al-Rfou, Rami and Sapp, Benjamin},
  booktitle={Proceedings of the IEEE/CVF International Conference on Computer Vision},
  pages={8579--8590},
  year={2023}
}

@misc{huang2025drivegptscalingautoregressivebehavior,
      title={DriveGPT: Scaling Autoregressive Behavior Models for Driving}, 
      author={Xin Huang and Eric M. Wolff and Paul Vernaza and Tung Phan-Minh and Hongge Chen and David S. Hayden and Mark Edmonds and Brian Pierce and Xinxin Chen and Pratik Elias Jacob and Xiaobai Chen and Chingiz Tairbekov and Pratik Agarwal and Tianshi Gao and Yuning Chai and Siddhartha Srinivasa},
      year={2025},
      eprint={2412.14415},
      archivePrefix={arXiv},
      primaryClass={cs.LG},
      url={https://arxiv.org/abs/2412.14415}, 
}

@misc{xu2024drivegpt4interpretableendtoendautonomous,
      title={DriveGPT4: Interpretable End-to-end Autonomous Driving via Large Language Model}, 
      author={Zhenhua Xu and Yujia Zhang and Enze Xie and Zhen Zhao and Yong Guo and Kwan-Yee. K. Wong and Zhenguo Li and Hengshuang Zhao},
      year={2024},
      eprint={2310.01412},
      archivePrefix={arXiv},
      primaryClass={cs.CV},
      url={https://arxiv.org/abs/2310.01412}, 
}

@misc{feng2025artemisautoregressiveendtoendtrajectory,
      title={ARTEMIS: Autoregressive End-to-End Trajectory Planning with Mixture of Experts for Autonomous Driving}, 
      author={Renju Feng and Ning Xi and Duanfeng Chu and Rukang Wang and Zejian Deng and Anzheng Wang and Liping Lu and Jinxiang Wang and Yanjun Huang},
      year={2025},
      eprint={2504.19580},
      archivePrefix={arXiv},
      primaryClass={cs.RO},
      url={https://arxiv.org/abs/2504.19580}, 
}

@INPROCEEDINGS{11093346,
  author={Zhang, Dongkun and Liang, Jiaming and Guo, Ke and Lu, Sha and Wang, Qi and Xiong, Rong and Miao, Zhenwei and Wang, Yue},
  booktitle={2025 IEEE/CVF Conference on Computer Vision and Pattern Recognition (CVPR)}, 
  title={CarPlanner: Consistent Auto-regressive Trajectory Planning for Large-scale Reinforcement Learning in Autonomous Driving}, 
  year={2025},
  volume={},
  number={},
  pages={17239-17248},
  doi={10.1109/CVPR52734.2025.01607}}

@misc{sun2024generalizingmotionplannersmixture,
      title={Generalizing Motion Planners with Mixture of Experts for Autonomous Driving}, 
      author={Qiao Sun and Huimin Wang and Jiahao Zhan and Fan Nie and Xin Wen and Leimeng Xu and Kun Zhan and Peng Jia and Xianpeng Lang and Hang Zhao},
      year={2024},
      eprint={2410.15774},
      archivePrefix={arXiv},
      primaryClass={cs.RO},
      url={https://arxiv.org/abs/2410.15774}, 
}

@misc{jiang2025transdiffuserdiversetrajectorygeneration,
      title={TransDiffuser: Diverse Trajectory Generation with Decorrelated Multi-modal Representation for End-to-end Autonomous Driving}, 
      author={Xuefeng Jiang and Yuan Ma and Pengxiang Li and Leimeng Xu and Xin Wen and Kun Zhan and Zhongpu Xia and Peng Jia and Xianpeng Lang and Sheng Sun},
      year={2025},
      eprint={2505.09315},
      archivePrefix={arXiv},
      primaryClass={cs.RO},
      url={https://arxiv.org/abs/2505.09315}, 
}

@article{li2024hydra,
  title={Hydra-mdp: End-to-end multimodal planning with multi-target hydra-distillation},
  author={Li, Zhenxin and Li, Kailin and Wang, Shihao and Lan, Shiyi and Yu, Zhiding and Ji, Yishen and Li, Zhiqi and Zhu, Ziyue and Kautz, Jan and Wu, Zuxuan and others},
  journal={arXiv preprint arXiv:2406.06978},
  year={2024}
}

@inproceedings{liao2025diffusiondrive,
  title={Diffusiondrive: Truncated diffusion model for end-to-end autonomous driving},
  author={Liao, Bencheng and Chen, Shaoyu and Yin, Haoran and Jiang, Bo and Wang, Cheng and Yan, Sixu and Zhang, Xinbang and Li, Xiangyu and Zhang, Ying and Zhang, Qian and others},
  booktitle={Proceedings of the Computer Vision and Pattern Recognition Conference},
  pages={12037--12047},
  year={2025}
}

@inproceedings{ICLR20256aa49679,
 author = {Li, Yingyan and Fan, Lue and He, Jiawei and Wang, Yuqi and Chen, Yuntao and Zhang, Zhaoxiang and Tan, Tieniu},
 booktitle = {International Conference on Representation Learning},
 editor = {Y. Yue and A. Garg and N. Peng and F. Sha and R. Yu},
 pages = {42942--42959},
 title = {Enhancing End-to-End Autonomous Driving with Latent World Model},
 url = {https://proceedings.iclr.cc/paper_files/paper/2025/file/6aa4967920e495e90aeeaa3acf18d019-Paper-Conference.pdf},
 volume = {2025},
 year = {2025}
}

@misc{cen2025worldvlaautoregressiveactionworld,
      title={WorldVLA: Towards Autoregressive Action World Model}, 
      author={Jun Cen and Chaohui Yu and Hangjie Yuan and Yuming Jiang and Siteng Huang and Jiayan Guo and Xin Li and Yibing Song and Hao Luo and Fan Wang and Deli Zhao and Hao Chen},
      year={2025},
      eprint={2506.21539},
      archivePrefix={arXiv},
      primaryClass={cs.RO},
      url={https://arxiv.org/abs/2506.21539}, 
}

@misc{li2025unifiedvideoactionmodel,
      title={Unified Video Action Model}, 
      author={Shuang Li and Yihuai Gao and Dorsa Sadigh and Shuran Song},
      year={2025},
      eprint={2503.00200},
      archivePrefix={arXiv},
      primaryClass={cs.RO},
      url={https://arxiv.org/abs/2503.00200}, 
}

@inproceedings{NIPS2017_7a98af17,
 author = {van den Oord, Aaron and Vinyals, Oriol and kavukcuoglu, koray},
 booktitle = {Advances in Neural Information Processing Systems},
 editor = {I. Guyon and U. Von Luxburg and S. Bengio and H. Wallach and R. Fergus and S. Vishwanathan and R. Garnett},
 pages = {},
 publisher = {Curran Associates, Inc.},
 title = {Neural Discrete Representation Learning},
 url = {https://proceedings.neurips.cc/paper_files/paper/2017/file/7a98af17e63a0ac09ce2e96d03992fbc-Paper.pdf},
 volume = {30},
 year = {2017}
}

@inproceedings{lu2023imitation,
  title={Imitation is not enough: Robustifying imitation with reinforcement learning for challenging driving scenarios},
  author={Lu, Yiren and Fu, Justin and Tucker, George and Pan, Xinlei and Bronstein, Eli and Roelofs, Rebecca and Sapp, Benjamin and White, Brandyn and Faust, Aleksandra and Whiteson, Shimon and others},
  booktitle={2023 IEEE/RSJ International Conference on Intelligent Robots and Systems (IROS)},
  pages={7553--7560},
  year={2023},
  organization={IEEE}
}

@inproceedings{hu2023planning,
  title={Planning-oriented autonomous driving},
  author={Hu, Yihan and Yang, Jiazhi and Chen, Li and Li, Keyu and Sima, Chonghao and Zhu, Xizhou and Chai, Siqi and Du, Senyao and Lin, Tianwei and Wang, Wenhai and others},
  booktitle={Proceedings of the IEEE/CVF conference on computer vision and pattern recognition},
  pages={17853--17862},
  year={2023}
}

@article{chen2024vadv2,
  title={Vadv2: End-to-end vectorized autonomous driving via probabilistic planning},
  author={Chen, Shaoyu and Jiang, Bo and Gao, Hao and Liao, Bencheng and Xu, Qing and Zhang, Qian and Huang, Chang and Liu, Wenyu and Wang, Xinggang},
  journal={arXiv preprint arXiv:2402.13243},
  year={2024}
}

@article{zhou2025autovla,
  title={AutoVLA: A Vision-Language-Action Model for End-to-End Autonomous Driving with Adaptive Reasoning and Reinforcement Fine-Tuning},
  author={Zhou, Zewei and Cai, Tianhui and Zhao, Seth Z and Zhang, Yun and Huang, Zhiyu and Zhou, Bolei and Ma, Jiaqi},
  journal={arXiv preprint arXiv:2506.13757},
  year={2025}
}

@misc{jiao2025evadriveevolutionaryadversarialpolicy,
      title={EvaDrive: Evolutionary Adversarial Policy Optimization for End-to-End Autonomous Driving}, 
      author={Siwen Jiao and Kangan Qian and Hao Ye and Yang Zhong and Ziang Luo and Sicong Jiang and Zilin Huang and Yangyi Fang and Jinyu Miao and Zheng Fu and Yunlong Wang and Kun Jiang and Diange Yang and Rui Fan and Baoyun Peng},
      year={2025},
      eprint={2508.09158},
      archivePrefix={arXiv},
      primaryClass={cs.LG},
      url={https://arxiv.org/abs/2508.09158}
}

@article{hwang2024emma,
  title={Emma: End-to-end multimodal model for autonomous driving},
  author={Hwang, Jyh-Jing and Xu, Runsheng and Lin, Hubert and Hung, Wei-Chih and Ji, Jingwei and Choi, Kristy and Huang, Di and He, Tong and Covington, Paul and Sapp, Benjamin and others},
  journal={arXiv preprint arXiv:2410.23262},
  year={2024}
}

@inproceedings{shao2023safety,
  title={Safety-enhanced autonomous driving using interpretable sensor fusion transformer},
  author={Shao, Hao and Wang, Letian and Chen, Ruobing and Li, Hongsheng and Liu, Yu},
  booktitle={Conference on Robot Learning},
  pages={726--737},
  year={2023},
  organization={PMLR}
}

@article{shao2024deepseekmath,
  title={Deepseekmath: Pushing the limits of mathematical reasoning in open language models},
  author={Shao, Zhihong and Wang, Peiyi and Zhu, Qihao and Xu, Runxin and Song, Junxiao and Bi, Xiao and Zhang, Haowei and Zhang, Mingchuan and Li, YK and Wu, Yang and others},
  journal={arXiv preprint arXiv:2402.03300},
  year={2024}
}

@article{cheng2023adversarial,
  title={Adversarial preference optimization},
  author={Cheng, Pengyu and Yang, Yifan and Li, Jian and Dai, Yong and Du, Nan},
  journal={CoRR},
  year={2023}
}

@misc{li2025recogdrivereinforcedcognitiveframework,
      title={ReCogDrive: A Reinforced Cognitive Framework for End-to-End Autonomous Driving}, 
      author={Yongkang Li and Kaixin Xiong and Xiangyu Guo and Fang Li and Sixu Yan and Gangwei Xu and Lijun Zhou and Long Chen and Haiyang Sun and Bing Wang and Kun Ma and Guang Chen and Hangjun Ye and Wenyu Liu and Xinggang Wang},
      year={2025},
      eprint={2506.08052},
      archivePrefix={arXiv},
      primaryClass={cs.CV},
      url={https://arxiv.org/abs/2506.08052}, 
}

@INPROCEEDINGS{9811576,
  author={Vitelli, Matt and Chang, Yan and Ye, Yawei and Ferreira, Ana and Wołczyk, Maciej and Osiński, Błażej and Niendorf, Moritz and Grimmett, Hugo and Huang, Qiangui and Jain, Ashesh and Ondruska, Peter},
  booktitle={2022 International Conference on Robotics and Automation (ICRA)}, 
  title={SafetyNet: Safe Planning for Real-World Self-Driving Vehicles Using Machine-Learned Policies}, 
  year={2022},
  volume={},
  number={},
  pages={897-904},
  doi={10.1109/ICRA46639.2022.9811576}}

@article{schulman2014motion,
  title={Motion planning with sequential convex optimization and convex collision checking},
  author={Schulman, John and Duan, Yan and Ho, Jonathan and Lee, Alex and Awwal, Ibrahim and Bradlow, Henry and Pan, Jia and Patil, Sachin and Goldberg, Ken and Abbeel, Pieter},
  journal={The International Journal of Robotics Research},
  volume={33},
  number={9},
  pages={1251--1270},
  year={2014},
  publisher={Sage Publications Sage UK: London, England}
}

@misc{wang2025unifiedvisionlanguageactionmodel,
      title={Unified Vision-Language-Action Model}, 
      author={Yuqi Wang and Xinghang Li and Wenxuan Wang and Junbo Zhang and Yingyan Li and Yuntao Chen and Xinlong Wang and Zhaoxiang Zhang},
      year={2025},
      eprint={2506.19850},
      archivePrefix={arXiv},
      primaryClass={cs.CV},
      url={https://arxiv.org/abs/2506.19850}, 
}

@inproceedings{zhang2025bridging,
  title={Bridging past and future: End-to-end autonomous driving with historical prediction and planning},
  author={Zhang, Bozhou and Song, Nan and Jin, Xin and Zhang, Li},
  booktitle={Proceedings of the Computer Vision and Pattern Recognition Conference},
  pages={6854--6863},
  year={2025}
}

@inproceedings{jiang2023vad,
  title={Vad: Vectorized scene representation for efficient autonomous driving},
  author={Jiang, Bo and Chen, Shaoyu and Xu, Qing and Liao, Bencheng and Chen, Jiajie and Zhou, Helong and Zhang, Qian and Liu, Wenyu and Huang, Chang and Wang, Xinggang},
  booktitle={Proceedings of the IEEE/CVF International Conference on Computer Vision},
  pages={8340--8350},
  year={2023}
}

@article{yang2025less,
  title={Less is More: Lean yet Powerful Vision-Language Model for Autonomous Driving},
  author={Yang, Sheng and Zhan, Tong and Chen, Guancheng and Lu, Yanfeng and Wang, Jian},
  journal={arXiv preprint arXiv:2510.00060},
  year={2025}
}

@inproceedings{sun2025sparsedrive,
  title={Sparsedrive: End-to-end autonomous driving via sparse scene representation},
  author={Sun, Wenchao and Lin, Xuewu and Shi, Yining and Zhang, Chuang and Wu, Haoran and Zheng, Sifa},
  booktitle={2025 IEEE International Conference on Robotics and Automation (ICRA)},
  pages={8795--8801},
  year={2025},
  organization={IEEE}
}

@article{zhou2025opendrivevla,
  title={Opendrivevla: Towards end-to-end autonomous driving with large vision language action model},
  author={Zhou, Xingcheng and Han, Xuyuan and Yang, Feng and Ma, Yunpu and Knoll, Alois C},
  journal={arXiv preprint arXiv:2503.23463},
  year={2025}
}

@article{li2024does,
  title={Does end-to-end autonomous driving really need perception tasks?},
  author={Li, Peidong and Cui, Dixiao},
  journal={arXiv e-prints},
  pages={arXiv--2409},
  year={2024}
}

@article{renz2022plant,
  title={Plant: Explainable planning transformers via object-level representations},
  author={Renz, Katrin and Chitta, Kashyap and Mercea, Otniel-Bogdan and Koepke, A and Akata, Zeynep and Geiger, Andreas},
  journal={arXiv preprint arXiv:2210.14222},
  year={2022}
}

@inproceedings{dauner2023parting,
  title={Parting with misconceptions about learning-based vehicle motion planning},
  author={Dauner, Daniel and Hallgarten, Marcel and Geiger, Andreas and Chitta, Kashyap},
  booktitle={Conference on Robot Learning},
  pages={1268--1281},
  year={2023},
  organization={PMLR}
}

@inproceedings{cheng2024rethinking,
  title={Rethinking imitation-based planners for autonomous driving},
  author={Cheng, Jie and Chen, Yingbing and Mei, Xiaodong and Yang, Bowen and Li, Bo and Liu, Ming},
  booktitle={2024 IEEE International Conference on Robotics and Automation (ICRA)},
  pages={14123--14130},
  year={2024},
  organization={IEEE}
}

@inproceedings{huang2024dtpp,
  title={Dtpp: Differentiable joint conditional prediction and cost evaluation for tree policy planning in autonomous driving},
  author={Huang, Zhiyu and Karkus, Peter and Ivanovic, Boris and Chen, Yuxiao and Pavone, Marco and Lv, Chen},
  booktitle={2024 IEEE International Conference on Robotics and Automation (ICRA)},
  pages={6806--6812},
  year={2024},
  organization={IEEE}
}

@article{chitta2022transfuser,
  title={Transfuser: Imitation with transformer-based sensor fusion for autonomous driving},
  author={Chitta, Kashyap and Prakash, Aditya and Jaeger, Bernhard and Yu, Zehao and Renz, Katrin and Geiger, Andreas},
  journal={IEEE transactions on pattern analysis and machine intelligence},
  volume={45},
  number={11},
  pages={12878--12895},
  year={2022},
  publisher={IEEE}
}

@article{li2025end,
  title={End-to-end driving with online trajectory evaluation via bev world model},
  author={Li, Yingyan and Wang, Yuqi and Liu, Yang and He, Jiawei and Fan, Lue and Zhang, Zhaoxiang},
  journal={arXiv preprint arXiv:2504.01941},
  year={2025}
}

@article{li2025drivevla,
  title={DriveVLA-W0: World Models Amplify Data Scaling Law in Autonomous Driving},
  author={Li, Yingyan and Shang, Shuyao and Liu, Weisong and Zhan, Bing and Wang, Haochen and Wang, Yuqi and Chen, Yuntao and Wang, Xiaoman and An, Yasong and Tang, Chufeng and others},
  journal={arXiv preprint arXiv:2510.12796},
  year={2025}
}

@article{dauner2024navsim,
  title={Navsim: Data-driven non-reactive autonomous vehicle simulation and benchmarking},
  author={Dauner, Daniel and Hallgarten, Marcel and Li, Tianyu and Weng, Xinshuo and Huang, Zhiyu and Yang, Zetong and Li, Hongyang and Gilitschenski, Igor and Ivanovic, Boris and Pavone, Marco and others},
  journal={Advances in Neural Information Processing Systems},
  volume={37},
  pages={28706--28719},
  year={2024}
}

@article{caesar2021nuplan,
  title={nuplan: A closed-loop ml-based planning benchmark for autonomous vehicles},
  author={Caesar, Holger and Kabzan, Juraj and Tan, Kok Seang and Fong, Whye Kit and Wolff, Eric and Lang, Alex and Fletcher, Luke and Beijbom, Oscar and Omari, Sammy},
  journal={arXiv preprint arXiv:2106.11810},
  year={2021}
}

@inproceedings{caesar2020nuscenes,
  title={nuscenes: A multimodal dataset for autonomous driving},
  author={Caesar, Holger and Bankiti, Varun and Lang, Alex H and Vora, Sourabh and Liong, Venice Erin and Xu, Qiang and Krishnan, Anush and Pan, Yu and Baldan, Giancarlo and Beijbom, Oscar},
  booktitle={Proceedings of the IEEE/CVF conference on computer vision and pattern recognition},
  pages={11621--11631},
  year={2020}
}

@article{baniodeh2025scaling,
  title={Scaling Laws of Motion Forecasting and Planning--A Technical Report},
  author={Baniodeh, Mustafa and Goel, Kratarth and Ettinger, Scott and Fuertes, Carlos and Seff, Ari and Shen, Tim and Gulino, Cole and Yang, Chenjie and Jerfel, Ghassen and Choe, Dokook and others},
  journal={arXiv preprint arXiv:2506.08228},
  year={2025}
}

@misc{kaplan2020scalinglawsneurallanguage,
      title={Scaling Laws for Neural Language Models}, 
      author={Jared Kaplan and Sam McCandlish and Tom Henighan and Tom B. Brown and Benjamin Chess and Rewon Child and Scott Gray and Alec Radford and Jeffrey Wu and Dario Amodei},
      year={2020},
      eprint={2001.08361},
      archivePrefix={arXiv},
      primaryClass={cs.LG},
      url={https://arxiv.org/abs/2001.08361}, 
}

@misc{hoffmann2022trainingcomputeoptimallargelanguage,
      title={Training Compute-Optimal Large Language Models}, 
      author={Jordan Hoffmann and Sebastian Borgeaud and Arthur Mensch and Elena Buchatskaya and Trevor Cai and Eliza Rutherford and Diego de Las Casas and Lisa Anne Hendricks and Johannes Welbl and Aidan Clark and Tom Hennigan and Eric Noland and Katie Millican and George van den Driessche and Bogdan Damoc and Aurelia Guy and Simon Osindero and Karen Simonyan and Erich Elsen and Jack W. Rae and Oriol Vinyals and Laurent Sifre},
      year={2022},
      eprint={2203.15556},
      archivePrefix={arXiv},
      primaryClass={cs.CL},
      url={https://arxiv.org/abs/2203.15556}, 
}

@misc{wu2025generatingmultimodaldrivingscenes,
      title={Generating Multimodal Driving Scenes via Next-Scene Prediction}, 
      author={Yanhao Wu and Haoyang Zhang and Tianwei Lin and Lichao Huang and Shujie Luo and Rui Wu and Congpei Qiu and Wei Ke and Tong Zhang},
      year={2025},
      eprint={2503.14945},
      archivePrefix={arXiv},
      primaryClass={cs.CV},
      url={https://arxiv.org/abs/2503.14945}, 
}

@misc{cao2025pseudosimulationautonomousdriving,
      title={Pseudo-Simulation for Autonomous Driving}, 
      author={Wei Cao and Marcel Hallgarten and Tianyu Li and Daniel Dauner and Xunjiang Gu and Caojun Wang and Yakov Miron and Marco Aiello and Hongyang Li and Igor Gilitschenski and Boris Ivanovic and Marco Pavone and Andreas Geiger and Kashyap Chitta},
      year={2025},
      eprint={2506.04218},
      archivePrefix={arXiv},
      primaryClass={cs.RO},
      url={https://arxiv.org/abs/2506.04218}, 
}

@misc{yao2025drivesuprimprecisetrajectoryselection,
      title={DriveSuprim: Towards Precise Trajectory Selection for End-to-End Planning}, 
      author={Wenhao Yao and Zhenxin Li and Shiyi Lan and Zi Wang and Xinglong Sun and Jose M. Alvarez and Zuxuan Wu},
      year={2025},
      eprint={2506.06659},
      archivePrefix={arXiv},
      primaryClass={cs.RO},
      url={https://arxiv.org/abs/2506.06659}, 
}

@misc{wang2026meanfuserfastonestepmultimodal,
      title={MeanFuser: Fast One-Step Multi-Modal Trajectory Generation and Adaptive Reconstruction via MeanFlow for End-to-End Autonomous Driving}, 
      author={Junli Wang and Xueyi Liu and Yinan Zheng and Zebing Xing and Pengfei Li and Guang Li and Kun Ma and Guang Chen and Hangjun Ye and Zhongpu Xia and Long Chen and Qichao Zhang},
      year={2026},
      eprint={2602.20060},
      archivePrefix={arXiv},
      primaryClass={cs.CV},
      url={https://arxiv.org/abs/2602.20060}, 
}

\end{document}